\documentclass[12pt]{article}

\usepackage[utf8]{inputenc}
\usepackage{amsmath,amsfonts, amsthm}
\usepackage[a4paper, total={16cm, 26cm}]{geometry}
\usepackage[small]{titlesec}
\usepackage{graphicx,color}
\usepackage{mathrsfs}

\usepackage{algpseudocode,algorithm,algorithmicx}

\usepackage[font={small},labelfont={small}]{caption}

\title{\large \bf Confidence intervals for the random forest generalization error}

\author{
{\normalsize Paulo C. Marques F.} \\
\textit{\small Insper Institute of Education and Research, São Paulo, Brazil} \\
{\tt\footnotesize PauloCMF1@insper.edu.br}
}

\date{\footnotesize December 2021}

\begin{document}

\maketitle

\begin{abstract}
We show that the byproducts of the standard training process of a random forest yield not only the well known and almost computationally free out-of-bag point estimate of the model generalization error, but also give a direct path to compute confidence intervals for the generalization error which avoids processes of data splitting and model retraining. Besides the low computational cost involved in their construction, these confidence intervals are shown through simulations to have good coverage and appropriate shrinking rate of their width in terms of the training sample size.
\end{abstract}


\section{Introduction}

How confident can we be in the generalization capacity of a predictive model? Of the many devices discussed in the statistical learning literature \cite{esl,murphy,bishop}, a simple random split of the original data into training and test sets, and methods of folded cross-validation, stand out as the most common tools used to tackle the generalization issue. Availability of point estimates for the generalization error given by these procedures naturally raises the question of how to quantify the confidence involved in these estimates spending a manageable computational cost.

Random forests \cite{breimanRF,cutler,biau,athey} elegantly provide an alternative low cost (almost free) point estimate of the generalization error without requiring splittings of the data, and avoiding the computational burden of retraining the predictive model several times. The bagging mechanism \cite{breimanBAG} used to construct the ensemble of trees implies that each training data point is not used (stays ``out-of-bag'') when growing approximately $36.8\%$ of the trees in the forest. This property gives us the so called out-of-bag estimate of the random forest generalization error: for each observation, using a suitable loss function, we compute the predictive error made by the random subforest whose trees did not include the observation under consideration in its training process; the out-of-bag estimate is the average of these prediction errors over the whole training sample.

In this paper we develop confidence intervals for the random forest generalization error based on its out-of bag estimate, without entailing any processes of data splitting or model retraining. The idea is to treat the original training data plus the bookkeeping of the out-of bag process and the predictions made for the training set by each tree in the ensemble as an augmented training sample, which is bootstrapped to produce confidence intervals of specified nominal levels.

A general description of random forests and the definitions of the predictive metrics necessary for our arguments are given in Section \ref{rf}. Section \ref{boot} discusses the bootstrapping of the augmented training sample and the derived confidence intervals for the generalization error. Use of simulated datasets in Section \ref{sim} allows us to investigate the effective coverage of the proposed confidence intervals for different nominal levels and training sample sizes. Two classic machine learning datasets and two more recent datasets of house prices and consumer churn information are analysed in Section \ref{real}. Section \ref{library} gives pointers to the open source library \texttt{rangerror} \cite{github} written in R \cite{R} which implements the procedures described in the paper.

\section{Random forests and the generalization error}\label{rf}

We have a data generating process for the exchangeable sequence
$$
  (X,Y),(X_1,Y_1),(X_2,Y_2),\dots
$$
such that its pairs, given some distribution function $F$ living in a potentially large nonparametric family, are conditionally independent and identically distributed, each of them having distribution $F$. Each sequence element assumes values in $\mathbb{R}^p\times\mathscr{Y}$, in such a way that in regression contexts $\mathscr{Y}$ is the real line, while in classification problems $\mathscr{Y}$ is a set of class labels $\{1,2,\dots,L\}$. In what follows, to simplify our discussion, we begin by considering the regression case, subsequently describing at the end of the section how the results apply in the classification setting.

The Classification and Regression Trees (CART) algorithm, developed in the 1980's by Breiman et al. \cite{cart}, recursively partitions the space of predictor variables, greedily looking for splits in the training data which minimize a specified cost function.

Suppose that we resample a training set of size $n$, drawing $n$ observations uniformly with replacement, and that using this bootstrap sample \cite{efron} we train a tall regression tree using the CART algorithm. Repeating this process $B$ times, we have the regression trees $\hat{\psi}^{(1)},\dots,\hat{\psi}^{(B)}$, which are averaged to produce the aggregated regression function
$$
  \hat{\psi}_n(\;\cdot\;) = \frac{1}{B} \sum_{j=1}^B\hat{\psi}^{(j)}(\;\cdot\;).
$$
This general aggregation process of regression functions trained from bootstrap samples, known as bagging \cite{breimanBAG}, was further optimized by Breiman by uniformly drawing without replacement a random subset of $\texttt{mtry}\leq p$ predictors when deciding each split of each regression tree in the ensemble, resulting in the definition of a random forest \cite{breimanRF}.

From a bias-variance trade-off perspective \cite{esl}, the good predictive performance of random forests would come from the variance reduction provided by averaging tall regression trees -- which by construction have low bias and high variance -- and the additional breaking up of the correlations between predictions made by each tree in the ensemble provided by Breiman's randomized split decision mechanism.

For a random forest $\hat{\psi}_n$, its generalization error is defined as the expected prediction error
$$
  \gamma_n = \mathbb{E}\!\left[\left(Y-\hat{\psi}_n(X)\right)^2\right].
$$
It is important to notice that in this expectation the forest $\hat{\psi}_n$ is a random object viewed as a functional of the random training sample $\{(X_i,Y_i)\}_{i=1}^n$.

Random forests are special in the sense that the byproducts of their training process give us a way to estimate $\gamma_n$ directly. It is easy to prove that the use of bootstrap samples in the construction of a random forest implies that each training observation is not used when growing approximately $e^{-1}\approx36.8\%$ of the trees in the ensemble. Letting $\mathcal{O}_i\subset\{1,2,\dots,B\}$ denote the indexes of the trees for which the $i$-th training observation stayed out of the corresponding bootstrap samples (using Breiman's terminology, stayed ``out-of-bag''), the regression trees $\{\hat{\psi}^{(j)}:j\in\mathcal{O}_i\}$ form a random subforest for which we can treat the $i$-th training observation as a test data point, giving us the out-of-bag estimate of the generalization error as
$$
  \hat{\gamma}_n=\frac{1}{n}\sum_{i=1}^n\left(y_i-\frac{1}{|\mathcal{O}_i|}\sum_{j\in\mathcal{O}_i}\hat{\psi}^{(j)}(x_i)\right)^2,
$$
in which $|\mathcal{O}_i|$ denotes the number of trees whose indexes are in $\mathcal{O}_i$.

When we move to classification problems, an analogous process gives us the classification trees $\hat{\psi}^{(1)},\dots,\hat{\psi}^{(B)}$, whose predictions are aggregated by majority voting, yielding the random forest
$$
  \hat{\psi}_n(\,\cdot\,)=\text{Mode}\,\{\hat{\psi}^{(j)}(\,\cdot\,):j=1,\dots,B\}.
$$
In this classification context, the generalization error is defined as the expected prediction error
$$
  \gamma_n = \mathbb{E}\!\left[\mathbb{I}(Y\ne\hat{\psi}_n(X))\right],
$$
in which $\mathbb{I}(\,\cdot\,)$ is an indicator function; and the out-of-bag estimate of the generalization error is given by
$$
  \hat{\gamma}_n=\frac{1}{n}\sum_{i=1}^n\mathbb{I}(y_i\ne\text{Mode}\,\{\hat{\psi}^{(j)}(x_i):j\in\mathcal{O}_i\}).
$$

\section{Bootstrapping the augmented training sample}\label{boot}

Defining $\hat{y}_{ij}=\hat{\psi}^{(j)}(x_i)$, for $i=1,\dots,n$ and $j=1,\dots,B$, the idea to construct a confidence interval for the generalization error $\gamma_n$ based on its out-of-bag estimate $\hat{\gamma}_n$ is to observe that $\hat{\gamma}_n$ can be formally seen as a statistic of the augmented training sample
$$
  \mathscr{A}_n = \{(x_i,y_i,\mathcal{O}_i,\{\hat{y}_{ij}\}_{j=1}^B)\}_{i=1}^n,
$$
which includes the original training sample, the bookkeeping of the out-of-bag information resulting from the random forest training process, and the individual predictions made by each tree of the ensemble for each training observation.

Therefore, a confidence interval for the random forest generalization error $\gamma_n$ can be obtained as follows. For $m=1,\dots,M$, construct $\mathscr{A}_n^{*(m)}$ by uniformly sampling with replacement $n$ points from the augmented training sample $\mathscr{A}_n$, and after that compute the corresponding out-of-bag estimate $\hat{\gamma}_n^{*(m)}$. We obtain an approximate level $1-\alpha$ confidence interval for the generalization error $\gamma_n$ from the empirical $\alpha/2$ and $1-\alpha/2$ percentiles of $\{\hat{\gamma}_n^{*(1)},\dots,\hat{\gamma}_n^{*(M)}\}$.

The computation of the just described confidence interval for $\gamma_n$ is inexpensive. First, the random forest is trained only once with all the available data, producing the augmented training sample $\mathscr{A}_n$. Second, the computation of each $\hat{\gamma}_n^{*(m)}$ can be done efficiently by directly resampling $\{\hat{\gamma}_{(i)}\}_{i=1}^n$,
in which
$$
  \hat{\gamma}_{(i)} = \left(y_i-\frac{1}{|\mathcal{O}_i|}\sum_{j\in\mathcal{O}_i}\hat{y}_{ij}\right)^2
$$
or
$$
  \hat{\gamma}_{(i)} = \mathbb{I}(y_i\ne\text{Mode}\,\{\hat{y}_{ij}:j\in\mathcal{O}_i\}),
$$
in regression and classification contexts, respectively. The pseudocode of the procedure is given in Algorithm \ref{alg:ci}.

\begin{algorithm}[t]
  \caption{Confidence interval for the generalization error\label{alg:ci}}
  \begin{algorithmic}[1]
    \Require{Nominal confidence level $1-\alpha$, augmented training sample $\mathscr{A}_n=\{(x_i,y_i,\mathcal{O}_i,\{\hat{y}_{ij}\}_{j=1}^B)\}_{i=1}^n$, and number of bootstrap replications $M$.}
    \Statex
    \Function{ConfidenceInterval}{$\alpha$, $\mathscr{A}_n$, $M$}
      \If {\textrm{Regression}}
        \For{$i \gets 1 \textrm{ to } n$}
          \State $\hat{\gamma}_{(i)} \gets (y_i-|\mathcal{O}_i|^{-1}\sum_{j\in\mathcal{O}_i}\hat{y}_{ij})^2$
        \EndFor
      \ElsIf {\textrm{Classification}}
        \For{$i \gets 1 \textrm{ to } n$}
          \State $\hat{\gamma}_{(i)} \gets \mathbb{I}(y_i\ne\text{Mode}\,\{\hat{y}_{ij}:j\in\mathcal{O}_i\})$
        \EndFor
      \EndIf
      \For{$m \gets 1 \textrm{ to } M$}
        \State $\textsc{boot} \gets$ size $n$ uniform sample with replacement from $\{1,\dots,n\}$
        \State $\hat{\gamma}_n^{*(m)} \gets (1/n) \sum_{i\in\textsc{BOOT}} \hat{\gamma}_{(i)}$
      \EndFor
      \State \Return{$\alpha/2$ and $1-\alpha/2$ percentiles of $\{\hat{\gamma}_n^{*(1)},\dots,\hat{\gamma}_n^{*(M)}\}$}
    \EndFunction
  \end{algorithmic}
\end{algorithm}

The advantage of this bootstrap procedure over a computation based on a central limit theorem for the average of the $\hat{\gamma}_{(i)}$'s is that the confidence intervals obtained from percentiles of the bootstrapped out-of-bag estimates are automatically invariant by monotonic transformations of the quantity of interest, which comes handy specially in regression problems such as those discussed in Sections \ref{subsec:autompg} and \ref{subsec:ames}, since it allows us to easily produce confidence intervals in units which are more natural for the interpretation of the results. Also, in classification problems we know that $\gamma_n\in[0,1]$ and the bootstrap procedure is guaranteed to produce confidence intervals which cover only valid values of the generalization error.

\section{Simulated data and coverage}\label{sim}

In this section we investigate the effective coverage of the proposed confidence intervals for the generalization error $\gamma_n$ using simulated regression and classification datasets.

The general Monte Carlo simulation procedure goes as follows. A training sample of size $n$ is simulated, from which a random forest $\hat{\psi}_n$ is grown. The augmented training sample $\mathscr{A}_n$ is a direct byproduct of the training procedure used to grow $\hat{\psi}_n$. A confidence interval for $\gamma_n$ with nominal level $1-\alpha$ is obtained from $\mathscr{A}_n$ as described in Algorithm \ref{alg:ci}. A very large test sample of size $n_\text{tst}$ is simulated from which a test error is computed using the predictions made by $\hat{\psi}_n$. The whole procedure is replicated $N$ times, and the fraction of replications in which the test errors stay inside the corresponding confidence interval gives us an approximation of the effective coverage. We also record the average over Monte Carlo replications of the computed confidence intervals widths.

\subsection{Regression}

In our regression example we simulate the data from the Friedman process discussed in \cite{friedman}.

We have ten independent predictors $X_1,\dots,X_{10}$, each of them having distribution $\text{U}[0,1]$, and a random variable $\epsilon$ having a standard normal distribution, independent of the $X_j$'s. The response variable is defined as
$$
  Y = 10 \sin (\pi X_1 X_2) + 20 (X_3 -1/2)^2 + 10 X_4 + 5 X_5 + \epsilon,
$$
implying that the response is not related to the last five predictors, which act as noise in the data.

The simulation results are presented in Table \ref{tab:sims} for two training sample sizes, showing that the simulated coverages are close to the corresponding nominal confidence levels.

\begin{table}[t!]
\centering
\caption{Monte Carlo results for the simulated datasets Friedman (regression) and Gaussian spheres (classification), doing $N=10^3$ replications. The number of trees in each random forest is $B=10^3$, the test sample size used to approximate the true generalization error is $n_\text{tst}=10^5$. We bootstrapped the augmented training sample $M=10^3$ times.}
\footnotesize
\begin{tabular}{ccccccccc}
\cline{2-9}
\multicolumn{1}{l}{} & \multicolumn{8}{c}{\textbf{Dataset}} \\ \cline{2-9} 
\multicolumn{1}{l}{} & \multicolumn{4}{c|}{\textbf{Friedman}} & \multicolumn{4}{c}{\textbf{Spheres}} \\ \cline{2-9} 
\multicolumn{1}{l}{} & \multicolumn{2}{c|}{$n = 500$} & \multicolumn{2}{c|}{$n = 1000$} & \multicolumn{2}{c|}{$n = 500$} & \multicolumn{2}{c}{$n = 1000$} \\ \hline
\multicolumn{1}{c|}{\textbf{nominal}} & \multicolumn{1}{l|}{\textbf{coverage}} & \multicolumn{1}{l|}{\textbf{avg len}} & \multicolumn{1}{l|}{\textbf{coverage}} & \multicolumn{1}{l|}{\textbf{avg len}} & \multicolumn{1}{l|}{\textbf{coverage}} & \multicolumn{1}{l|}{\textbf{avg len}} & \multicolumn{1}{l|}{\textbf{coverage}} & \multicolumn{1}{l}{\textbf{avg len}} \\ \hline
0.05 & 0.039 & 0.04105 & 0.056 & 0.02400 & 0.031 & 0.00086 & 0.044 & 0.00061 \\
0.10 & 0.091 & 0.08242 & 0.100 & 0.04816 & 0.111 & 0.00262 & 0.093 & 0.00180 \\
0.15 & 0.137 & 0.12379 & 0.140 & 0.07276 & 0.152 & 0.00384 & 0.134 & 0.00254 \\
0.20 & 0.183 & 0.16566 & 0.182 & 0.09750 & 0.188 & 0.00457 & 0.186 & 0.00354 \\
0.25 & 0.222 & 0.20819 & 0.234 & 0.12252 & 0.252 & 0.00624 & 0.223 & 0.00431 \\
0.30 & 0.262 & 0.25171 & 0.276 & 0.14834 & 0.314 & 0.00760 & 0.278 & 0.00534 \\
0.35 & 0.312 & 0.29628 & 0.325 & 0.17468 & 0.351 & 0.00857 & 0.330 & 0.00619 \\
0.40 & 0.366 & 0.34183 & 0.373 & 0.20196 & 0.418 & 0.01024 & 0.388 & 0.00721 \\
0.45 & 0.410 & 0.38974 & 0.425 & 0.22997 & 0.470 & 0.01161 & 0.424 & 0.00816 \\
0.50 & 0.459 & 0.43972 & 0.480 & 0.25941 & 0.525 & 0.01286 & 0.478 & 0.00925 \\
0.55 & 0.512 & 0.49256 & 0.523 & 0.29023 & 0.571 & 0.01467 & 0.525 & 0.01033 \\
0.60 & 0.563 & 0.54841 & 0.573 & 0.32342 & 0.631 & 0.01623 & 0.574 & 0.01154 \\
0.65 & 0.626 & 0.60879 & 0.622 & 0.35903 & 0.668 & 0.01808 & 0.624 & 0.01279 \\
0.70 & 0.673 & 0.67518 & 0.672 & 0.39803 & 0.718 & 0.01999 & 0.693 & 0.01419 \\
0.75 & 0.717 & 0.74887 & 0.722 & 0.44194 & 0.754 & 0.02220 & 0.749 & 0.01576 \\
0.80 & 0.756 & 0.83479 & 0.776 & 0.49212 & 0.813 & 0.02469 & 0.786 & 0.01754 \\
0.85 & 0.795 & 0.93708 & 0.825 & 0.55226 & 0.847 & 0.02771 & 0.849 & 0.01970 \\
0.90 & 0.863 & 1.07089 & 0.893 & 0.63059 & 0.896 & 0.03164 & 0.893 & 0.02247 \\
0.95 & 0.920 & 1.27538 & 0.943 & 0.75016 & 0.939 & 0.03769 & 0.944 & 0.02675 \\
\hline
\end{tabular}
\label{tab:sims}
\end{table}

\subsection{Classification}

For the classification case we simulate the data from a slightly modified version of the Gaussian spheres example discussed on page 339 of \cite{esl}.

We have twenty independent predictors $X_1,\dots,X_{20}$, each of them having a standard normal distribution. The last ten predictors are not related to the response and behave like noise in the data. Define
$$
  Z = \begin{cases}
    \hfill 1 &\text{if} \quad \sum_{j=1}^{10} X_j^2 > \chi^2_{10}(0.5) \\
    \hfill -1 &\text{otherwise}
  \end{cases}
$$
in which $\chi^2_{10}(0.5)$ is the median of a chi-squared distribution with ten degrees of freedom. Letting $W$ be a random variable with $\text{Bernoulli}(0.05)$ distribution, independent of the $X_j$'s, we use $W$ to randomly flip the sign of $Z$, defining the two-class response variable $Y=(-1)^W\cdot Z$.

The simulation results are presented in Table \ref{tab:sims} for a range of nominal confidence levels and two training sample sizes. There we can see that the simulated coverages are again close to the specified nominal confidence levels.

Finally, inspection of Table \ref{tab:sims} indicates that the average over the Monte Carlo replications of the confidence intervals widths are shrinking at an almost standard rate $O(n^{-0.498})$ for the Gaussian spheres problem, and at the faster rate $O(n^{-0.764})$ for the Friedman process.

\section{Real datasets}\label{real}

In this section we compute confidence intervals for the generalization error of random forests trained on two classic machine learning datasets from the UCI repository \cite{UCI}, and two more recent datasets of housing prices and consumer churn behavior. At the end of the section we discuss the running time of Algorithm \ref{alg:ci} for the four datasets.

\subsection{Auto MPG}\label{subsec:autompg}

The Auto MPG dataset \cite{quinlan} from the UCI repository \cite{UCI} contains information about the city-cycle fuel consumption of $392$ cars. There are $8$ predictor variables, and the goal in this regression task is to predict the fuel consumption in miles per gallon. Table \ref{tab:classic} gives the confidence intervals for the generalization error at different confidence levels. We used the invariance by monotonic transformations property of the confidence intervals produced by Algorithm \ref{alg:ci} to report the lower and upper limits of the intervals in miles per gallon.

\subsection{Spam data}\label{subsec:spam}

The Spambase dataset available at the UCI repository \cite{UCI} contains information about $4601$ e-mails which were classified as legitimate (``ham'') or illegitimate (``spam''). We have $57$ predictor variables describing the percentages of occurrences of certain words and lengths of capitalization patterns appearing in the messages. The confidence intervals for the generalization error are presented for different levels in Table \ref{tab:classic}.

\subsection{Ames housing}\label{subsec:ames}

The housing dataset discussed in \cite{decock} contains information about $2930$ houses sold in Ames, Iowa from 2006 to 2010. We have eighty predictors and the response variable is the sale price in US dollars. Table \ref{tab:recent} gives the confidence intervals for the generalization error at different confidence levels. In this regression task, we used again the invariance by monotonic transformations property of the confidence intervals produced by Algorithm \ref{alg:ci} to report the lower and upper limits of the intervals in US dollars.

\subsection{Telecom churn}

The dataset described in \cite{jafari} contains information about $3150$ clients of a Telecommunications company. We have thirteen predictors and the response variable indicates if the clients cancelled the service (churned) or not. Table \ref{tab:recent} gives the confidence intervals.

\subsection{Running time}

We studied the time required in a standard notebook equiped with a $4.60$ GHz Intel processor $\textrm{i7-8565U}$ to compute the confidence intervals for the four real datasets discussed above. Table \ref{tab:timing} presents descriptive summaries of the computation times in miliseconds based on $100$ replications of Algorithm \ref{alg:ci} for each of the four datasets, always using $M=10^3$ bootstrap replications.

\begin{table}[t!]
\centering
\caption{Confidence intervals for two classic UCI \cite{UCI} datasets at different confidence levels. The augmented training sample was bootstrapped $M=10^3$ times.}
\small
\begin{tabular}{ccccc}
\cline{2-5}
 & \multicolumn{4}{c}{\textbf{Dataset}} \\ \cline{2-5} 
 & \multicolumn{2}{c|}{\textbf{Auto MPG} ($n = 392$)} & \multicolumn{2}{c}{\textbf{Spam} ($n = 4601$)} \\ \hline
\multicolumn{1}{c|}{\textbf{confidence level}} & \multicolumn{1}{c|}{\;\;\;\;\;\;\textbf{lower}\;\;\;\;\;\;} & \multicolumn{1}{c|}{\textbf{upper}} & \multicolumn{1}{c|}{\;\;\;\;\;\;\textbf{lower}\;\;\;\;\;\;} & \textbf{upper} \\ \hline
0.90 & 2.41 & 3.02 & 0.0417 & 0.0517 \\
0.95 & 2.35 & 3.09 & 0.0411 & 0.0537 \\
0.99 & 2.33 & 3.22 & 0.0380 & 0.0554 \\
\hline
\end{tabular}
\label{tab:classic}
\end{table}

\begin{table}[t!]
\centering
\caption{Confidence intervals for two real datasets at different confidence levels. The augmented training sample was bootstrapped $M=10^3$ times.}
\small
\begin{tabular}{ccccc}
\cline{2-5}
 & \multicolumn{4}{c}{\textbf{Dataset}} \\ \cline{2-5} 
 & \multicolumn{2}{c|}{\textbf{Ames housing} ($n = 2930$)} & \multicolumn{2}{c}{\textbf{Telecom churn} ($n = 3150$)} \\ \hline
\multicolumn{1}{c|}{\textbf{confidence level}} & \multicolumn{1}{c|}{\;\;\;\;\;\;\textbf{lower}\;\;\;\;\;\;} & \multicolumn{1}{c|}{\textbf{upper}} & \multicolumn{1}{c|}{\;\;\;\;\;\;\textbf{lower}\;\;\;\;\;\;} & \textbf{upper} \\ \hline
0.90 & 23,907.11 & 27,418.47 & 0.0352 & 0.0467 \\
0.95 & 23,657.64 & 27,788.31 & 0.0346 & 0.0476 \\
0.99 & 23,165.04 & 28,505.73 & 0.0314 & 0.0505 \\
\hline
\end{tabular}
\label{tab:recent}
\end{table}

\begin{table}[t!]
\centering
\caption{Descriptive summaries for the running times in miliseconds based on $100$ replications of Algorithm \ref{alg:ci}. On each replication the augmented training sample was bootstrapped $M=10^3$ times.}
\small
\begin{tabular}{lcccc}
\hline
\multicolumn{1}{l|}{\textbf{Dataset}} & \multicolumn{1}{c|}{\textbf{Training sample size}} & \multicolumn{1}{c|}{\textbf{median (ms)}} & \multicolumn{1}{c|}{\textbf{min (ms)}} & \textbf{max (ms)} \\ \hline
Auto MPG & 392 & 68.8 & 59.6 & 153.6 \\
Spam & 4601 & 1008.7 & 815.5 & 1241.5 \\
Ames housing & 2930 & 580.9 & 515.3 & 723.2 \\
Telecom churn & 3150 & 540.5 & 456.1 & 914.5 \\ \hline
\end{tabular}
\label{tab:timing}
\end{table}

\section{Open source software library}\label{library}

An open source R library \texttt{rangerror} implementing the procedures of the paper is available at \cite{github}, with instalation instructions and a few examples. It is based on the output of the impressive random forest library \texttt{ranger} \cite{wright}.

\bibliographystyle{ieeetr}

\bibliography{references}

\end{document}